\documentclass[num-refs]{wiley-article}

\usepackage{siunitx}
\usepackage{amsmath}
\usepackage{subfiles}

\papertype{Original Article}

\title{FatSegNet: A Fully Automated Deep Learning Pipeline for Adipose Tissue Segmentation on Abdominal Dixon MRI }

\author[1,2]{Santiago Estrada}
\author[2]{Ran Lu}
\author[1]{Sailesh Conjeti}
\author[2]{Ximena Orozco-Ruiz}
\author[2]{Joana Panos-Willuhn}
\author[2,3]{Monique M.B. Breteler}
\author[1,4,5]{Martin Reuter}

\affil[1]{Image Analysis, German Center for Neurodegenerative Diseases (DZNE), Bonn, Germany}
\affil[2]{Population Health Sciences, German Center for Neurodegenerative Diseases (DZNE), Bonn, Germany}
\affil[3]{Institute for Medical Biometry, Informatics and Epidemiology (IMBIE), Faculty of Medicine, University of Bonn, Bonn, Germany}
\affil[4]{A.A. Martinos Center for Biomedical Imaging, Massachusetts General Hospital, Boston MA, USA }
\affil[5]{Department of Radiology, Harvard Medical School, Boston MA,USA}

\corraddress{Santiago Estrada, DZNE, Venusberg-Campus 1 BLDG~99 , 53127 Bonn, Germany}
\corremail{santiago.estrada@dzne.de}

\fundinginfo{ Diet-BB Competence Cluster in Nutrition Research, Federal Ministry of Education and Research (BMBF), Grant Numbers: 01EA1410C and FKZ: 01EA1809C ; JPI HDHL on Biomarkers for Nutrition and Health,BMBF, Grant Number: 01EA1705B; National Institutes of Health (NIH),Grant Number: R01NS083534 and R01LM012719}
\runningauthor{Submitted to Magnetic Resonance in Medicine}
\begin{document}
\maketitle
\begin{abstract}
\textbf{Purpose:} Introduce and validate a novel, fast, and fully automated deep learning pipeline (FatSegNet) to accurately identify, segment, and quantify visceral and subcutaneous adipose tissue (VAT and SAT) within a consistent, anatomically defined abdominal region on Dixon MRI scans.

\noindent
\textbf{Method:} FatSegNet is composed of three stages: (i) Consistent localization of the abdominal region using two 2D-Competitive Dense Fully Convolutional Networks (CDF\-Net), (ii) Segmentation of adipose tissue on three views by independent CDFNets, and (iii) View-aggregation. FatSegNet is validated by: 1) comparison of segmentation accuracy (sixfold cross-validation), 2) test-retest reliability, 3) generalizability to randomly selected manually re-edited cases, and 4) replication of age and sex effects in the Rhineland Study - a large prospective population cohort.

\noindent
\textbf{Results:} The CDFNet demonstrates increased accuracy and robustness compared to traditional deep learning networks. FatSegNet Dice score outperforms manual raters on VAT (0.850~vs.~0.788),and produces comparable results on SAT (0.975~vs.~0.982). 
The pipeline has excellent agreement for both test-retest (ICC VAT 0.998 and SAT 0.996) and manual re-editing (ICC VAT 0.999 and SAT 0.999).

\noindent
\textbf{Conclusion:} FatSegNet generalizes well to different body shapes, sensitively replicates known VAT and SAT volume effects in a large cohort study, and permits localized analysis of fat compartments. Furthermore, it can reliably analyze a 3D Dixon MRI in $\sim$~1~min, providing an efficient and validated  pipeline for abdominal adipose tissue analysis in the Rhineland Study.

\end{abstract}

\section{Introduction}
The excess of body fat depots is an increasing major public health issue worldwide and an important risk factor for the development of metabolic disorders and reduced quality of life~\cite{padwal2016relationship,ng2014global}. While the body mass index (BMI) is a widely used indicator of adipose tissue accumulation in the body, it does not provide information on fat distribution~\cite{tomiyama2016misclassification} -- neither with respect to different fat tissue types nor with respect to deposit location. Different compartments of adipose tissue are associated with different physiopathological effects ~\cite{linge2018body,despres2012body}. Abdominal adipose tissue (AAT), composed of  subcutaneous and visceral adipose tissue (SAT and VAT), has long been associated with an increased risk of chronic cardiovascular diseases, glucose impairment, and dyslipidemia~\cite{kissebah1982relation,despres1990regional}. Recently, several studies have indicated a stronger relation between the accumulation of VAT with an adverse metabolic and inflammatory profile compared to SAT~\cite{despres2006abdominal,de2014visceral}. Therefore, an accurate and independent measurement of VAT and SAT volumes (VAT-V and SAT-V) is of significant clinical and research interest.

Currently, the gold standard for measuring VAT-V and SAT-V is the manual segmentation of abdominal fat images from Dixon magnetic resonance (MR) scans -- a very expensive and time-consuming process. Thus, especially for large studies, automatic segmentation methods are requi\-red. However, achieving good accuracy is challenging due to complex AAT structures, a wide variety of VAT shapes, large anatomical differences across subjects, and the inherent properties of the Dixon images: low intensity contrast between adipose tissue classes, inhomogeneous signals, and potential organ motion. So far, those limitations impeded the wide-spread implementation of automatic and semi-automatic techniques based on intensity and shape features, such as fuzzy-clustering~\cite{zhou2011novel}, k-means clustering~\cite{thormer2013software}, graph cut~\cite{christensen2017automatic,sadananthan2015automated} active contour methods~\cite{mosbech2011automatic}, and statistical shape models~\cite{wald2012automatic}. 

Recently, fully convolutional neural networks (F-CNNs) \cite{segnet,FCNNS} have been widely adopted in the computer vision community for pixel/voxel-wise image segmentation in an end-to-end fashion to overcome above-mentioned challenges. With these methods there is no need to extract manual features, divide images into patches, or implement sliding window techniques. F-CNNs  can automatically extract intrinsic features and integrate global context to resolve local ambiguities thereby improving the results of the predicted models~\cite{FCNNS}. Langer~\textit{et al.}~\cite{adiposefully} proposed a three channel UNet for AAT segmentation, which is a conventional architecture for 2D medical image segmentation~\cite{unet}. While this method showed promising results, we demonstrate that our network architecture outperforms the traditional UNet for segmenting AAT on our images with a wide range of anatomical variation.
 More recent architectures such as the SD-Net~\cite{sd-net} and Dense-UNet, a densely connected network~\cite{Densenet}, have the potential to improve generalizability and robustness by encouraging feature re-usability and strengthening information propagation across the network~\cite{Densenet}. In prior work, we introduced a competitive dense fully convolutional network (CDFNet)~\cite{concatvscomp} as a new 2D F-CNN architecture that promotes feature selectivity within a network by introducing maximum attention through a maxout activation unit~\cite{maxout}. The maxout boosts performance by allowing the creation of specialized sub-networks that target a specific structure during training~\cite{competitivenet}. Therefore, this approach facilitates the learning of more complex structures~\cite{concatvscomp,competitivenet} with the added benefit of reducing the number of training parameters relative to the aforementioned networks. 

In this paper, we propose FatSegNet, a novel fully automated deep learning pipeline based on our CDFNet architecture to localize and segment VAT and SAT on abdominal Dixon MR images from the Rhineland Study, an ongoing large population-based cohort study~\cite{breteler2014mri,stocker2016big}. To constrain AAT segmentations to a consistent anatomically defined region, the proposed pipeline consists of three stages: 
\begin{enumerate}
    \item {\bf Localization} of the abdominal region using a semantic segmentation approach by implementing CDFNet models on sagittal and coronal planes; we use the lumbar vertebrae positions as reference points for selecting the region of interest.
    \item {\bf Segmentation} of VAT and SAT within the abdominal region through 2D CDFNet models on three different planes (axial, sagittal and coronal).
     \item A {\bf view-aggregation} stage where the previous generated label maps are combined to generate a final 3D segmentation. 
\end{enumerate}

We initially evaluate and compare the individual stages of the pipeline with other deep learning approaches in a sixfold cross-validation. We show that the proposed network architecture (CDFNet) improves segmentation performance and simultaneously reduces the number of required training parameters in step 1 and 2. After asserting segmentation accuracy, we evaluate the whole pipeline (FatSegNet) with respect to robustness and reliability against two independent test sets: a manually edited and a test-retest set. Finally, we present a case study on unseen data comparing the VAT-V and SAT-V calculated from the FatSegNet segmentations against BMI to replicate age and sex effects on these volumes in a large cohort.

\section{Methods}
\subsection{Data}
\subsubsection{MR imaging acquisition}
 MR image acquisition was performed at two different sites both with identical 3T Siemens MAGNETOM Prisma MR scanners (Siemens Healthcare, Erlangen, Germany). The body coil was used for signal reception of a three-dimen\-sional two-point Dixon sequence (acquisition time =12~$s$, echo time TE1=1.23~$ms$, TE2=2.46~$ms$, repetition time TR=4.12~$ms$, axial field of view =500~$mm \times$ 437~$mm$, flip angle =6~degrees, left-right readout bandwidth =750~$Hz\-/pixel$, partial Fourier factor 6/8 $\times$ 5/8). Based on a preceding moving-table abdominal localizer, the field-of-view was centered on the middle of the third lumbar vertebra (L, L3). Data were acquired during a single breath-hold in supine position with arms placed at the sides. The image resolution was finally interpolated from 2.0~$mm \times$ 2.7~$mm \times$ 10.0~$mm$ to 2.0~$mm \times$ 2.0~$mm \times$ 5.0~$mm$ (matrix size =256 $\times$ 224 $\times$ 72).
 
\begin{figure*}[!hbt]
\centering
\includegraphics[width=\textwidth]{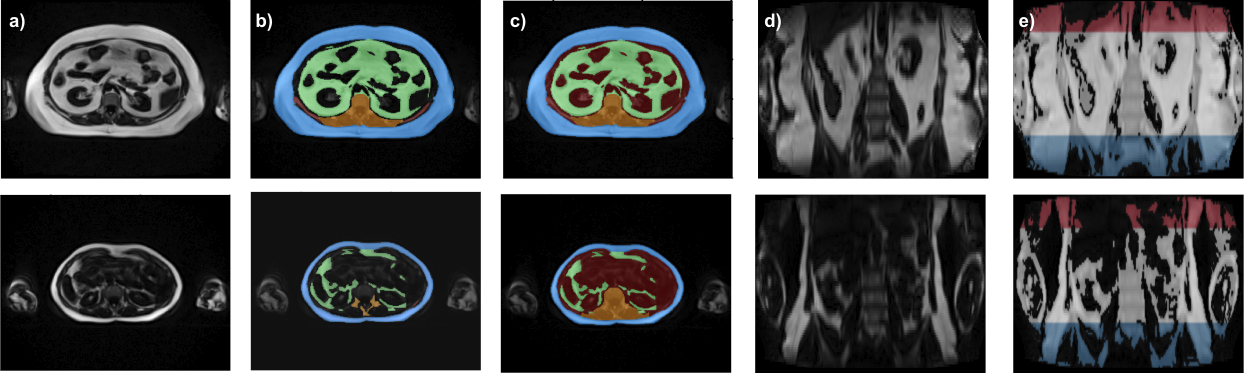}
\caption{\textbf{MR Dixon images and ground truth from two subjects with different BMI (obese~(upper), normal~(lower)}. a) Fat images: axial plane. b) Initial manual segmentation (blue: SAT, green: VAT, orange: bone and surrounding structures). c) Ground truth with additional synthetic class (red: other-tissue) and filled-in bone structures (orange). d) Fat images: coronal plane. e) Ground truth for localization of region of interest (red: thoracic region, white: abdominal region (region of interest), blue: pelvic region). }
\label{fig:ground_Truth}
\end{figure*}

\subsubsection{Datasets}
 The Rhineland Study is an ongoing population-based pro\-spective cohort~(https://www.rheinland-studie.de/) which enrolls participants aged 30 years and above at baseline from Bonn, Germany. The study is carried out in accordance with the recommendations of the International Council for Harmonisation (ICH) Good Clinical Practice (GCP) standards (ICH-GCP). Written informed consent was obtained from all participants in accordance with the Declaration of Helsinki. \\
 The first 641 subjects from the Rhineland Study with BMI and abdominal MR Dixon scans are included. The sample presents a mean age of 54.2 years (range 30 to 95) and 55.2\% of the subjects are women. The BMI of the participants ranges from 17.2 to 47.7~kg/m$^2$ with a mean of 25.2~kg/m$^2$. Subjects were stratified into two subsets: 38 scans were manually annotated for training and testing; the remaining 603 subjects were segmented using the proposed pipeline. After visual inspection, 16 subjects were excluded due to poor image quality or extreme motion artifacts (e.g.\ potentially caused by breathing). Thus, 587 participants were used for the case study analysis and a subset of 50 subjects were randomly selected for manual corrections of the predicted label maps. This manually edited set and an independent test-retest set of 17 healthy young volunteers were used to assess reliability of the automated segmentation and volume estimates.

\noindent
 \textbf{Ground Truth Data}: 38 subjects were randomly selected from sex and BMI strata to ensure a balanced population distribution. These scans were manually annotated by two trained raters without any semi-automated support such as thresholding, which can reduce accuracy in the ground truth and lead to overestimation of the performance of the proposed automated method. 
 
 Specific label schemes were created for each individual task of the pipeline. For localizing the abdominal region, raters divided the scans into three different blocks defined by the location of the vertebrae as follows: the abdominal region (from lower bound of twelfth thoracic vertebra (Th12) to the lower bound of L5), the thoracic region (all above the lower bound of Th12), and the pelvic region (everything below the lower bound of L5), as illustrated in Fig.~\ref{fig:ground_Truth}~e). For AAT segmentation, 60 slices per subject were manually labeled into three classes: SAT, VAT, and bone with neighbouring tissues. The bone was labeled to prevent bone marrow from being misclassified as adipose tissue. In order to improve spatial context and prevent misclassification of the arms, the dataset was complemented by a synthetic class defined as ``other tissue'' that was composed of any soft tissue inside the abdomen cavity that is not VAT or SAT. The manual annotations are illustrated in Fig.~\ref{fig:ground_Truth} b) and c). Furthermore, four subjects were labeled by both raters to evaluate the inter-rater variability.
 
\noindent
\textbf{Test-Retest Data}: 17 additional subjects were recruited with the exclusive purpose of measuring the acquisition protocol reliability. The group presents a mean age of 25.5 years (range: 20 to 31) and 65.0~\% of the participants are women; all of them have a normal BMI (BMI < 25~kg/m$^2$). Subjects were scanned in two consecutive sessions. Before starting the second session, subjects were removed from the scanner and re-positioned.

\subsection{FatSegNet Pipeline}
\begin{figure*}[!hbt]
\centering
\includegraphics[width=\textwidth]{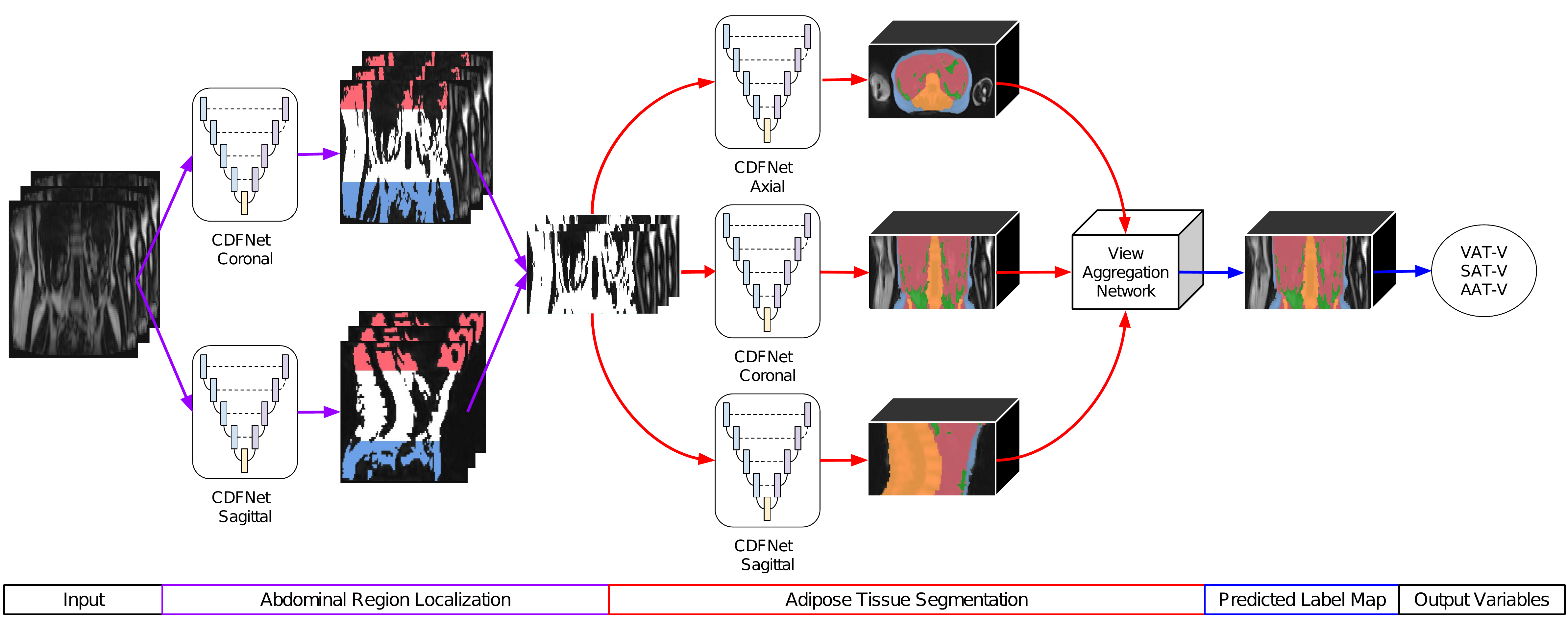}
\caption{\textbf{Proposed FatSegNet Pipeline for segmenting AAT}. The pipeline is divided into three stages: First, localization of abdominal region. Then, tissue segmentation on the abdominal region and finally, view-aggregation. Both local and global volume estimates of individual structures are calculated on the final prediction.}
\label{fig:pipeline}
\end{figure*}

The FatSegNet is to be deployed as a post-processing adipose analysis pipeline for the abdominal Dixon MR images acquired in the Rhineland Study. Therefore, it should meet the following requirements:
1) be fully automated,
2) segment the different adipose tissue types within the anatomically defined abdominal region, and 
3) be robust to body type variations and generalizable in presence of high population heterogeneity.
Following the prior conditions, we designed FatSegNet as a fully automated deep learning pipeline for adipose segmentation (Fig.~\ref{fig:pipeline}).

The proposed pipeline consists of three stages: (i)the abdominal region is localized by averaging bounding boxes from two abdominal segmentation maps generated by CDFNets on the sagittal and coronal view. For each view a bounding box is set to the full image width. The height is extracted by localizing the highest and lowest slice with at least 85\% of none background voxels classified as abdominal region. Highest and lowest slice position are averaged across the views. (ii) Afterward, adipose tissue is segmented within the abdominal region by three CDFNets on different views (axial, coronal, and sagittal) with standardized input sizes (zero padding). (iii) Finally, a view-aggregation network merges the predicted label maps from the previous stage into a final segmentation; the implemented multi-view scheme is designed to improve segmentation of structures that are not clearly visible due to poor lateral resolution. This 2.5D strategy produces a fully automated pipeline to accurately segment adipose tissue inside a consistent anatomically defined abdominal region.

\subsubsection{Pipeline components}

\begin{figure*}[!t]
\centering
\includegraphics[width=\textwidth]{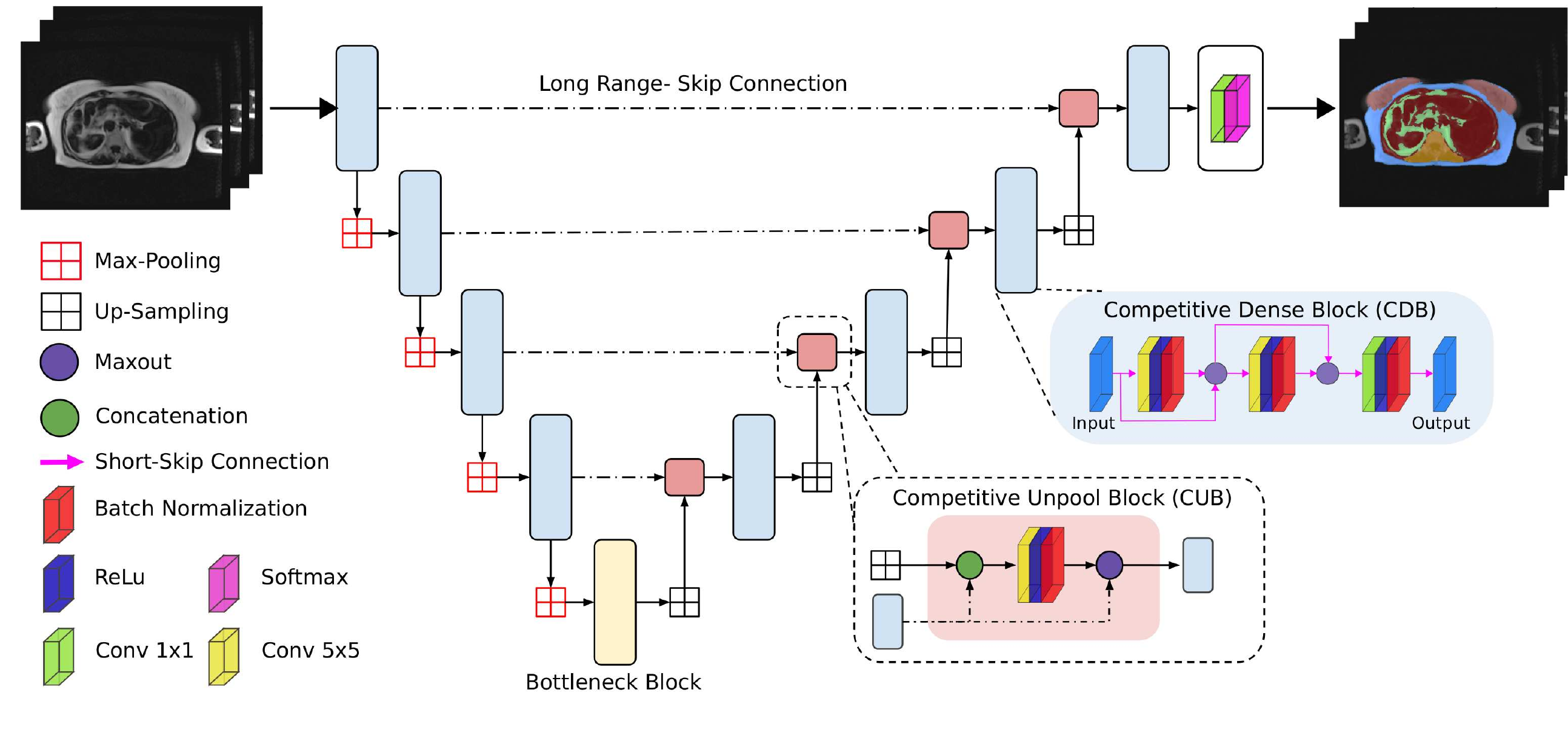}
\caption{\textbf{Proposed network architecture} : Competitive Dense Fully Convolutional Network (CDFNet), with 4 competitive dense blocks (CDB) on each encoder and decoder path and 4 competitive unpool blocks (CUB) between them. CDB and CUB induce local and global competition within the network. Note - the output filters for all convolutional layers in CUB, CDB, and Bottleneck were standardized to 64 channels.}
\label{fig:CDFNet}
\end{figure*}

\noindent \textbf{Competitive Dense Fully Convolutional Network  (CDF\-Net):}
For the segmentation task we introduce the CDF\-Net architecture due to its robustness and generalizability properties. The proposed network improves feature selectivity and, thus, boosts the learning of fine-grained anatomies without increasing the number of learned parameters~\cite{concatvscomp}. We implemented the CDFNet by suitably adopting the Dense-UNet architecture proposed by Roy~\textit{et al.}~\cite{quicknat} and extending it towards competitive learning via maxout activations~\cite{competitivenet}.

The Dense-UNet proposed in~\cite{quicknat} follows the usual dumb-bell like architecture with four dense-block encoders, four dense-block decoders and one bottleneck layer. Each dense-block is based on short-range-skip-connections between convolutional layers as introduced for densely-connected neural networks~\cite{dense_connections}; the dense connection approach stacks multiple convolutional layers in sequence and the input of a layer is iteratively concatenated with the outputs of the previous layers. This type of connectivity improves feature reusability, increases information propagation, and alleviates vanishing gradients~\cite{dense_connections}. The architecture additionally incorporates the traditional long-range skip-connec\-tions between all encoder and decoder blocks of the same spatial resolution as introduced by Ronnenberger~\textit{et al.}~\cite{unet} which improves gradient flow and spatial information recovery. 

Within the Dense-UNet, the information aggregation through these connections is performed by concatenation layers. Such a design increases the size of the output feature map along the feature channels, which in turn results in the need to learn filters with a higher number of parameters. Goodfellow~\textit{et al.} introduced the idea of competitive learning through maxout activations~\cite{maxout}, which was adapted by Liao and Carneiro~\cite{competitivenet} for competitive pooling of multi-scale filter outputs. Both~\cite{maxout} and~\cite{competitivenet} proved that the use of maxout competitive units boosts performance by creating a large number of dedicated sub-networks within a network that learns to target specific sub-tasks and reduces the number of required parameters significantly, which in turn can prevent over-fitting.

The maxout is a simple feed-forward activation function that chooses the maximum value from its inputs~\cite{maxout}. Within a CNN, a maxout feature map is constructed by taking the maximum across multiple input feature maps for a particular spatial location. The proposed CDFNet uses competitive layers (maxout activation) instead of concatenation layers. Our preliminary results~\cite{concatvscomp} demonstrate that these competitive units promote the formation of dedicated local sub-networks in each of the densely connected blocks within the encoder and the decoder paths. This encourages sub-modularity through a network-in-network design that can learn more efficiently. Towards this, we propose two novel architectural elements targeted at introducing competition within the short- and long-range connections, as follows: 
\begin{itemize}
    \item \textbf{Local Competition - Competitive Dense Block (CDB):}
     By introducing maxout activations within the short-range skip-connections of each of the densely connected convolutional layers (at the same resolution), we encourage local competition during learning of filters. The multiple convolution layers in each block prevent filter co-adaptation.
  
     \item \textbf{Global Competition - Competitive Un-pooling Block (CUB):}  We introduce a maxout activation between a long-range skip-connection from the encoder and the features up-sampled from the prior lower-resolution decoder block. This promotes competition between finer feature maps with smaller receptive fields (skip connections) and coarser feature maps from the decoder path that spans much wider receptive fields encompassing higher contextual information. 
\end{itemize}

In brief, the proposed CDFNet comprises a sequence of four CDBs, constituting the encoder path (down\--sampling block), and four CDBs constituting the decoder path (up\--sampling block), which is joined via a bottleneck layer. The bottleneck consists of a 2D convolutional layer followed by a Batch Normalization. The skip-connections from each of the encoder blocks feed into the CUB that subsequently forwards features into the corresponding decoder block of the same resolution as illustrated in Fig.~\ref{fig:CDFNet}. 

\noindent
\textbf{View-Aggregation Network}
The proposed view-aggregation network is designed to regularize the prediction for a given voxel by considering spatial information from the coronal, axial, and sagittal view. The network, therefore, merges the probability maps of the three different CDF\-Nets from the previous stage by applying a ($ 3 \times 3 \times 3$) 3D-convolution (30 filters) followed by a Batch Normalization. Then a ($1 \times 1 \times 1 $) 3D-convolution is employed to reduce the feature maps to the desired number of classes (n=5). The final prediction probabilities are obtained via a concluding softmax layer (as illustrated in supporting information Fig.~\ref{fig:view_Agg}). Our approach learns to weigh each view differently on a voxel level, compared to standard hard-coded global view-aggregation schemes. Such hard-coded weighting schemes can be suboptimal when working with anisotropic voxels sizes (e.g., here 2~$mm$ $\times$ 2~$mm$ $\times$ 5~$mm$) as resolution differences impose a challenge when combining the spatial information from the finer (within-plane) and coarser (across slice) resolutions. Additionally, in the presence of high variance of abdominal body shapes across subjects segmentation benefits from data-driven approaches that can flexibly adopt weights to individual situations and even spatial locations, which are not possible if hard-coded global weights are being used.

\subsection{Experimental setup} 
For training and testing the pipeline, we perform a sixfold cross-validation sub\-ject-space split on the ground truth dataset. For each fold, 32 subjects are used for training and 6 held out for testing; the test sets splits are approximately balanced based on their BMI classification (underweight [BMI<18.5~kg/m$^2$], normal [18.5 $\leq$ BMI<25~kg/m$^2$], overweight [25 $\leq$ BMI<30~kg/m$^2$], and obese [BMI $\geq$ 30~kg/m$^2$]). This selection process ensures that all BMI categories are used for bench-marking the cross-validation models. Additionally, a final model is implemented using 33 subjects for training holding out 5 subjects spanning different BMI levels for a final performance sanity-check (visual quality check and stability of Dice score). Given the limited ground truth data, for all models a validation set to assets convergence during training was created by randomly separating 15\% of the slices from the corresponding training set. This allows evaluating performance and generalizability on a completely separate test set.

\noindent
\textbf{Baselines and comparative methods:} We validate the FatSegNet by comparing the performance of each stage of the pipeline against the cross-validation test sets using Dice score index~(DSC) to measure similarity between the prediction and the ground truth. Let M (ground truth) and P (prediction) denote the labels binary segmentation, the Dice score index is defined as
\begin{eqnarray}
         DSC=  \frac{2\cdot \left | M \cap P\right | } { \left | M \right | + \left |  P\right|}
    \end{eqnarray}

Where $|M|$ and $|P|$ represents the number of elements in each segmentation, and $| M \cap P |$ the number of common elements. Therefore, the DSC ranges from 0 to 1 and a higher DSC represents a better agreement between segmentations.

Additionally, we benchmark the proposed CDFNet models for abdominal region localization and AAT delineation with state-of-the-art segmentation F-CNNs such as UNet~\cite{unet}, SD-Net~\cite{sd-net}, and Dense-UNet~\cite{quicknat}. We use the probability maps generated from the aforementioned networks to train the view-aggregation model and measure performance with and without view-aggregation. The proposed view-aggregation performance for each FCNNs is compared against two non-data-driven (hard-coded) methods: equally balanced weights for all views, and axial focus weights (accounting for higher in-plane resolution, axial=0.5, coronal=0.25, sagittal=0.25). Finally, to permit a fair comparison, all benchmark networks follow the same architecture of four encoder blocks, four decoders blocks, and one bottleneck layer as illustrated in Fig.~\ref{fig:CDFNet} with an input image size of 224$\times$ 256. Note, significant differences between our proposed methods and comparative baselines are evaluated by a Wilcoxon signed-rank test~\cite{wilcoxon1945individual} after multiple comparisons correction using a one-sided adaptive FDR~\cite{FDR2}.

The aforementioned models are implemented in Ke\-ras~\cite{chollet2015keras} with a TensorFlow back-end using an NVIDIA Titan Xp GPU with 12 GB RAM and the following parameters: batch size of 8, momentum set to 0.9, constant weight decay of 10$^{-06}$, and an initial learning rate of 0.01 decreased by a order of 10 every 20 epochs. The models are trained for 60 epochs with an early-stopping criterion (no relevant changes on the validation loss after the last 8 epochs -- convergence was observed around 50 epochs). A composite loss function of median frequency balanced logistic loss and Dice loss~\cite{sd-net} is used. This loss function emphasizes the boundaries between classes and supports learning of unbalanced classes such as VAT. Finally, online data augmentation (translation, rotation and global scaling) is performed to increase training set size and improve the networks generalizability. Note, the FatSegNet implementation is available at https://github.com/reuter-lab/FatSegNet.

\noindent
\textbf{Pipeline reliability:} We assess the FatSegNet reliability by comparing the difference of VAT-V and SAT-V across sessions for each subject of the test-retest and manually edited set. Given a predicted label map and $N_{i}(l)$ the number of voxels classified as $l$ (VAT or SAT) in session $i$ (test-retest, or manual-automated), the absolute percent difference (APD~($l$)) of a label volume measures variability across sessions. It is defined as
    \begin{eqnarray}
         APD(l)=  \frac{2\cdot \left | N_{1}(l)-N_{2}(l)\right | } { N_{1}(l) + N_{2}(l)} \cdot 100
     \end{eqnarray}
    
 Additionally, we calculate the agreement of total VAT-V and SAT-V between sessions by an intra-class correlation (ICC) using a two-way fixed, absolute agreement and single measures ICC(A,1)~\cite{ICC}.

\noindent
\textbf{Case study analysis on the Rhineland Study:} We compare the volumes of abdominal adipose tissue (AAT-V, SAT-V, and VAT-V) generated from FatSegNet with BMI on the unseen dataset. A fast quality control is performed to identify drastic failure cases. The differences among BMI groups are evaluated with a one-way analysis of variance (ANOVA) with subsequent Tukey's honest significant difference (HSD) post-hoc comparisons. The associations of volumes of abdominal adipose tissue and BMI are assessed using partial correlation and linear regression after accounting for age, sex, and height of the abdominal region. Separate linear regression analyses are performed to explore the effect of age on SAT-V and VAT-V in men and women. All the statistical analyses are performed in R~\cite{rsoftware}. 

\section{Results}

\subsection{Method Validation}

\noindent
\textbf{Localization of abdominal region}: For assessing the performance of abdominal region detection after creation of an average bounding box from the coronal and sagittal views the average Dice overlap (sixfold cross-validation) was calculated, as illustrated on the supporting information Fig.~\ref{fig:loc_boxplot}. We observe that all models perform extremely well on the relatively easy task of localizing the desired abdominal region (DSC > 0.96). There is no significant difference between the models; however, we use our CDFNet because it requires substantially less parameters~(see Table~\ref{table:AATsegmentation}) compared to the UNet and Dense-UNet. 

\noindent
\textbf{Segmentation of AAT}: In Table~\ref{table:AATsegmentation}, we present the average Dice score (sixfold cross-validation) for VAT and SAT for each individual view as well as for the view-aggregation model. Here, we observe that all methods work extremely well for SAT segmentation. Nevertheless, our proposed CDFNet outperforms the UNet and SD-Net on all single-view models and, when compared with the Dense-UNet, there is significant improvement in the sagittal and coronal views. For the more challenging task of VAT recognition, which is a more fine-grained compartment with large shape variation, the proposed CDFNet outperforms the SD-Net on all single planes; when compared with Dense-UNet and U-Net, there is only significant improvement in the axial and coronal plane. Nonetheless, CDFNet achieves this performance with $\sim$30~\%( Dense-UNet) and $\sim$80~\%(UNet) less parameters, demonstrating that the proposed architecture improves feature selectivity and simplifies network learning. Furthermore, fewer parameters can help decrease over-fitting error, especially when training with limited annotated data, and thus improve generalizability.

\begin{table*}[!hbt]
\centering
\caption{Mean (and standard deviation) Dice scores (cross-validation) of the FCNN models for abdominal adipose tissue segmentation. We show FDR corrected significance indicators of Wilcoxon signed-rank test~\cite{wilcoxon1945individual} comparing the proposed CDFNet vs. benchmark FCNNs}
\label{table:AATsegmentation}
\resizebox{\textwidth}{!}{
\begin{threeparttable}
\begin{tabular}{l|cccc|cccc}
\headrow
 &\multicolumn{4}{c}{\textbf{Subcutaneous (SAT)}} & \multicolumn{4}{c}{\textbf{Visceral (VAT)}} \\
\textbf{Models (PRM)}$^\dagger$ & \textbf{Axial} & \textbf{Coronal} & \textbf{Sagittal} &\textbf{V. Aggregation} & \textbf{Axial} & \textbf{Coronal} & \textbf{Sagittal} &\textbf{V. Aggregation}\\
UNet~($\sim$20~M)  & 0.965 (0.029)$^{\ast}$ & 0.960 (0.034)$^{\ast}$  & 0.960 (0.035)$^{\ast}$  & 0.972 (0.019)$^{\ast}$  & 0.810 (0.111)$^{\ast}$ & 0.804 (0.113)$^{\ast}$ & 0.820 (0.101)  & 0.837 (0.095)$^{\ast}$  \\
SD-Net~($\sim$~1,5M)   & 0.969 (0.027)$^{\ast}$ & 0.954 (0.040)$^{\ast}$ & 0.956 (0.034)$^{\ast}$ & 0.972 (0.020)$^{\ast}$ & 0.820 (0.097)$^{\ast}$  & 0.812 (0.099)$^{\ast}$ & 0.822 (0.091)$^{\ast}$ & 0.843 (0.081)$^{\ast}$  \\
Dense-UNet~($\sim$~3,3M)   & 0.972 (0.025)$^{\ast}$  & 0.959 (0.037)$^{\ast}$  & 0.963 (0.029)$^{\ast}$  & 0.975 (0.019)$^{\ast}$  & 0.824 (0.091)$^{\ast}$ & 0.814 (0.097)$^{\ast}$  & 0.827 (0.090)$^{\ast}$ & 0.847(0.080)$^{\ast}$  \\
\hline
\textbf{Proposed}~($\sim$2,5M)   & 0.970 (0.025)  & 0.966 (0.029) & 0.966 (0.027)  & \textbf{0.975}(0.018) & 0.826 (0.095) & 0.826 (0.085)  & 0.824 (0.092)  & \textbf{0.850}(0.076)  \\ \hline
\headrow
\textbf{Inter-rater variability}& \multicolumn{4}{c}{0.982 (0.018)} & \multicolumn{4}{c}{0.788 (0.060)} \\
\hline  
\end{tabular}
\begin{tablenotes}  
\large{
\item $^\dagger$ The approximately number of learn parameters reported is for the models without the View-Aggregation Network
\item $^{\ast}$ Statistical difference using a one-sided adaptive FDR  multiple comparison correction~\cite{FDR2} at a level of 0.05}
\end{tablenotes}
\end{threeparttable}}
\end{table*}

 Note, that Dice scores increase and difference of pairwise comparisons is slightly reduced after the view-aggregation (Table~\ref{table:AATsegmentation}), showing that this steps helps all individual networks to reach a better performance  by introducing spatial information from multiple views and regularizing the prediction maps. The proposed data-driven aggregation scheme outperforms (DSC) the hard-coded models for SAT and with statistically significance for VAT as shown in Table~\ref{table:view_agg_comparison}. Furthermore, learned weights are spatially varying and can adjust to subject-specific anatomy, which in turn can improve generalizability. We empirically observe that the aggregation model smoothes the label maps slightly, resulting in visually more appealing boundaries. It also significantly reduces the arms from being misclassified as adipose tissue which can otherwise be observed in different views, especially on overweight and obese subjects, where arms are located closer to the abdominal cavity, as seen supporting information Fig.~\ref{fig:arms}. 
 
 Finally it should be highlighted, that all single-view and the view-aggregation models achieve similarly excellent results on the SAT segmentation compared to inter-rater variability and outperform the manual raters for the more challenging VAT segmentation by a margin.
 
\begin{table*}[!hbt]
\centering
\caption{ Mean (and standard deviation) Dice scores (cross-validation) of  hard-coded balanced weights, hard-coded axial focus weights, and the proposed view-aggregation for abdominal adipose tissue segmentation. We show FDR corrected significance indicators of Wilcoxon signed-rank test~\cite{wilcoxon1945individual} comparing the proposed data-driven aggregation scheme vs. each hard-coded method.}
\label{table:view_agg_comparison}
\resizebox{0.9\textwidth}{!}{
\begin{threeparttable}
\begin{tabular}{l|ccc|ccc}
\headrow
 &\multicolumn{3}{c}{\textbf{Subcutaneous (SAT)}} & \multicolumn{3}{c}{\textbf{Visceral (VAT)}} \\
\textbf{Single-View Model} & \textbf{Balanced} & \textbf{Axial Focus} & \textbf{Proposed}  & \textbf{Balanced} & \textbf{Axial Focus} & \textbf{Proposed} \\
\hline
UNet & 0.970~(0.026) & 0.970~(0.026)&\textbf{0.972}~(0.019) & 0.830~(0.098)$^{\ast}$  & 0.829~(0.099)$^{\ast}$  & \textbf{0.837~}(0.095) \\
SD-Net & 0.970~(0.026)$^{\ast}$  & 0.972~(0.025)$^{\ast}$ &\textbf{0.972}~(0.020) & 0.839~(0.084)$^{\ast}$  & 0.838~(0.085)$^{\ast}$  &	\textbf{0.843}~(0.082)\\
Dense-UNet & 0.973~(0.025) & 0.974~(0.024)$^{\ast}$  & \textbf{0.975}~(0.019) & 0.841~(0.081)$^{\ast}$& 0.840~(0.082)$^{\ast}$&\textbf{0.847}~(0.080)\\
CDFNet & 0.972~(0.025)$^{\ast}$   & 0.973~(0.024) & \textbf{0.975}~(0.018) & 0.844~(0.077)$^{\ast}$  & 0.841~(0.080)$^{\ast}$  & \textbf{0.850}~(0.076)\\
\hline
\end{tabular}
\begin{tablenotes}  
\item $^{\ast}$ Statistical difference using a one-sided adaptive FDR multiple comparison correction~\cite{FDR2} at a level of 0.05
\end{tablenotes}
\end{threeparttable}}
\end{table*}

\noindent
\textbf{FatSegNet reliability}: Table~\ref{table:reliability} presents the reliability metrics evaluated on the test-retest and the manually edited test set. 
The proposed pipeline presents only a small absolute percent volume difference (APD) for VAT and SAT, and excellent agreement between the predicted and corrected segmentation maps. It must be noted, that APD is larger for both tissue types in the test-retest setting 
as it also includes variance from acquisition noise (e.g.\ motion artefacts, non-linearities based on different positioning) in addition to potential variances of the processing pipelines. Nevertheless, we observe excellent agreement (ICC) between sessions for the test-retest dataset for both adipose tissue types.

\begin{table}[!hbt]
\caption{Mean absolute percent difference (APD) and interclass correlation agreement (ICC(A,1)) for the volumes estimates of VAT and SAT across sessions of the manually edited and test-retest set.}
\centering
\label{table:reliability}
\resizebox{0.80\textwidth}{!}{
\begin{tabular}{l|cc|cc}
\headrow
 & \multicolumn{2}{c}{\textbf{Manually Edited Set}} & \multicolumn{2}{c}{\textbf{Test-Retest Set}}  \\
\textbf{Metric} & \textbf{SAT-V}& \textbf{VAT-V} & \textbf{SAT-V} & \textbf{VAT-V} \\
ICC [95\%~CI] &  0.999~[0.999 - 1.000] &  0.999~[0.994 - 0.999] & 0.996~[0.986 - 0.999] &  0.998~[0.995 - 0.999] \\
APD (SD)    & 0.149~\% (0.424)   & 1.398~\% (0.963) & 3.254~\% (2.524)   & 2.957~\% (2.600) \\  
\hline  
\end{tabular}}
\end{table}

   \begin{figure}[!hbt]
\centering
\includegraphics[width=0.85\textwidth]{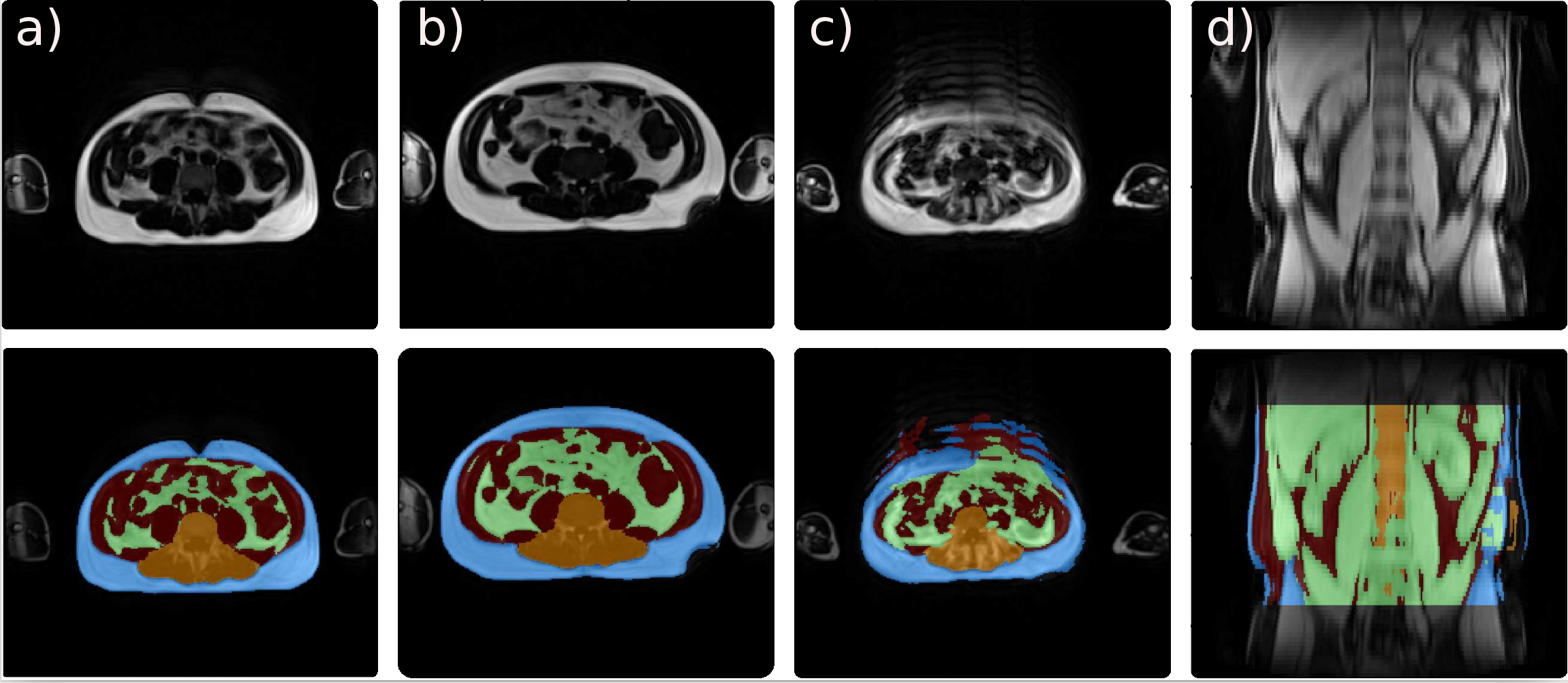}
\caption{\textbf{ Examples of FatSegNet predictions and excluded cases on the Rhineland Study.} (a-b) Subjects with different body shapes and accurate segmentations. (c-d) Excluded subjects from the case study due to extreme motion noise (c), or low image contrast quality (d).}
\label{fig:results_supp}
\end{figure}

\subsection{Case Study: Analysis of Rhineland Study Data}
\textbf{The characteristics of the study population:} After visual quality inspection, 16 scans were flagged due to image artefacts, such as motion or low contrast (see Fig.~\ref{fig:results_supp} c) and d) for two examples). 
The characteristics of the remaining 587 participants with valid data on BMI and volumes of abdominal adipose tissue are presented in supporting information Table~\ref{table:population}. The mean (SD) age of the subjects is 54.2 (13.3) years, and 54.7\% are women. 311 (53.0\%) subjects are normal weight, 209 (35.6\%) overweight, and 67 (11.4\%) obese. We observed a BMI increase with age ($\beta$ = 0.03, \textit{p}=0.007), and a borderline significance of age difference among BMI groups (\textit{p}=0.052,~ANOVA). Obvious differences are observed in AAT-V, VAT-V, and SAT-V across BMI groups (\textit{p}<0.001,~ANOVA). VAT-V to SAT-V ratio is higher in overweight and obese participants compared to those with normal weight (\textit{p}<0.001), but there is no difference between overweight and obese (\textit{p}=0.505). 

\begin{figure}[!b]
\centering
\includegraphics[width=0.5\columnwidth]{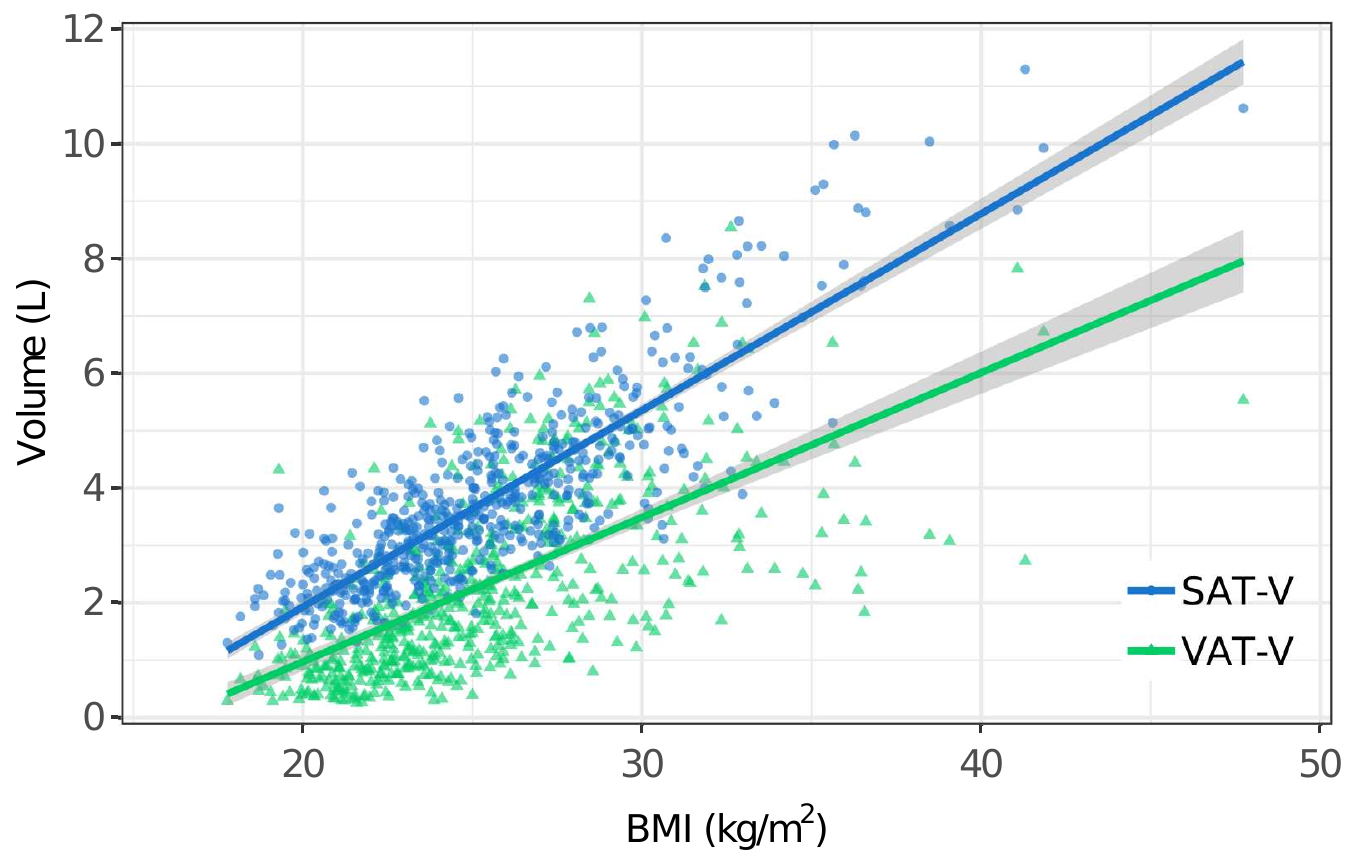}
\caption{\textbf{Association of BMI with SAT-Volume and VAT-Volume}}
\label{fig:bmivol}
\end{figure}

\noindent
\textbf{The association between abdominal adipose tissue volumes and BMI:}
BMI shows a strong positive correlation with AAT-V and SAT-V (AAT-V: r = 0.88, \textit{p}<0.001; SAT-V: r=0.85, \textit{p}<0.001), but only a moderate correlation with VAT-V (r=0.65, \textit{p}<0.001) after adjusting for age, sex, and abdominal region height. As illustrated in Fig.~\ref{fig:bmivol}, both SAT-V and VAT-V are positively associated with BMI after accounting for age, sex, and abdominal region height (\textit{p}<0.001). The accumulation of SAT-V is higher than VAT-V as BMI increases.

\begin{figure}[!hbt]
\centering
\includegraphics[width=0.7\columnwidth]{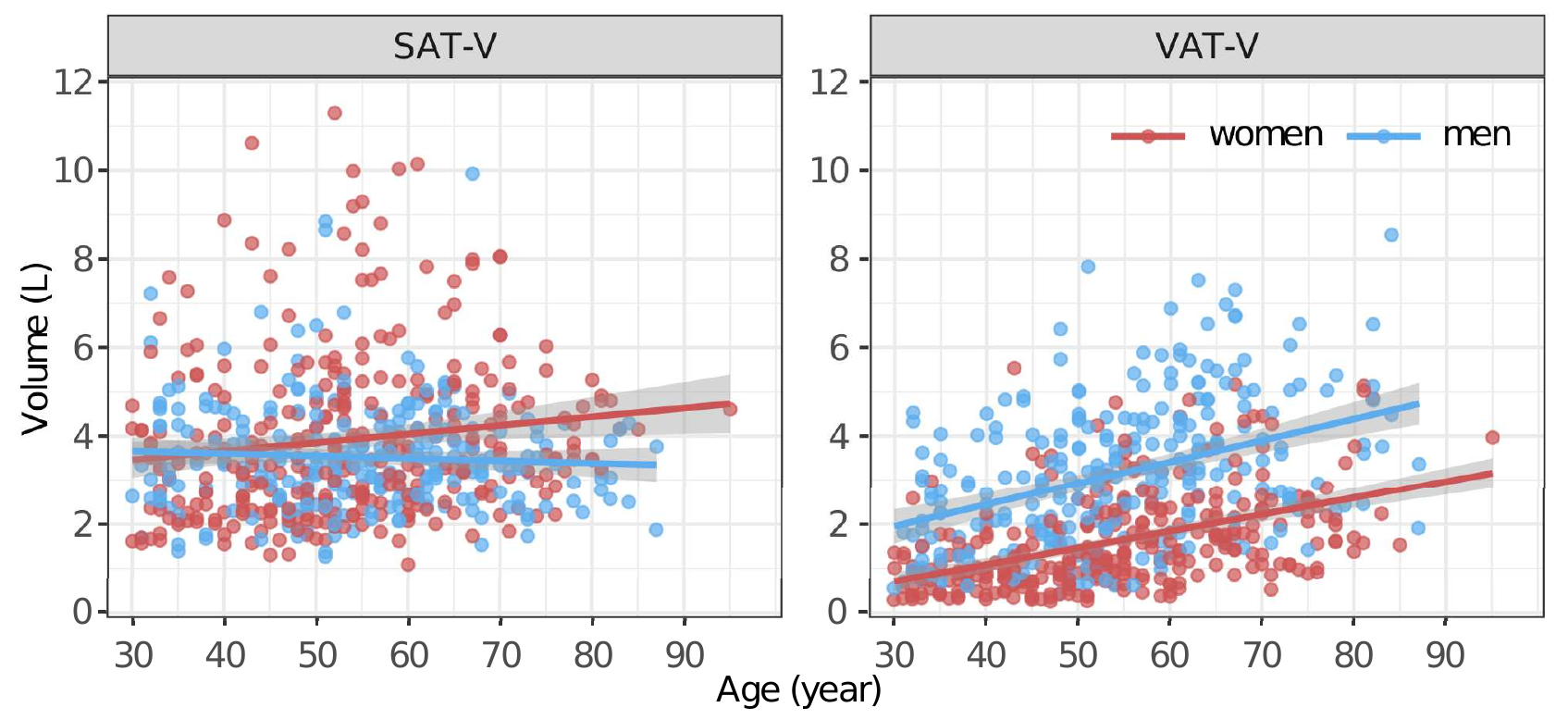}
\caption{\textbf{The association between age and SAT-Volume and VAT-Volume in men and women}.}
\label{fig:agesexvol}
\end{figure}

\noindent
\textbf{Influence of age and sex on VAT-V and SAT-V:}
The influence of age and sex on VAT-V and SAT-V follows different patterns (as illustrated in Fig.~\ref{fig:agesexvol}). Men tend to have lower SAT and higher VAT compared to women (\textit{p}<0.001). VAT-V significantly increase with age in both men and women. Conversely, SAT-V is weakly associated with age in women ($\beta$ = 0.02, \textit{p}=0.012), but not in men ($\beta$ = -0.01, \textit{p}=0.337). 

\section{Discussion}
In our study, we established, validated, and implemented a novel deep learning pipeline to segment and quantify the components of abdominal adipose tissue, namely, VAT-V, SAT-V, and AAT-V on a fast acquisition abdominal Dixon MR protocol for subjects from the Rhineland Study, a large population-based cohort. The proposed pipeline is fully automated and requires approximately 1~min for analyzing  a subject's whole volume. Moreover, since the pipeline is based on deep learning models, it can be easily updated and retrain\-ed as the study progresses and new manual data are generated - which can further improve overall pipe\-line robustness and generalizability, providing a pragmatic solution for a population-based study. 

The proposed pipeline, termed FatSegNet  implements a three-stage design with the CDFNet architecture at the core for localizing the abdominal region and segmenting the AAT. The introduction of our CDFNet inside the pipeline boosts the competition among filters to improve feature selectivity within the networks. CDFNet introduces competition at a local scale by substituting concatenation layers with maxout activations that prevent filter co-adapt\-ation and reduce the overall network complexity. It also induces competition at a global scale through competitive unpooling. This network design, in turn, can learn more efficiently.

\begin{figure*}[!t]
\centering
\includegraphics[width=0.8\textwidth]{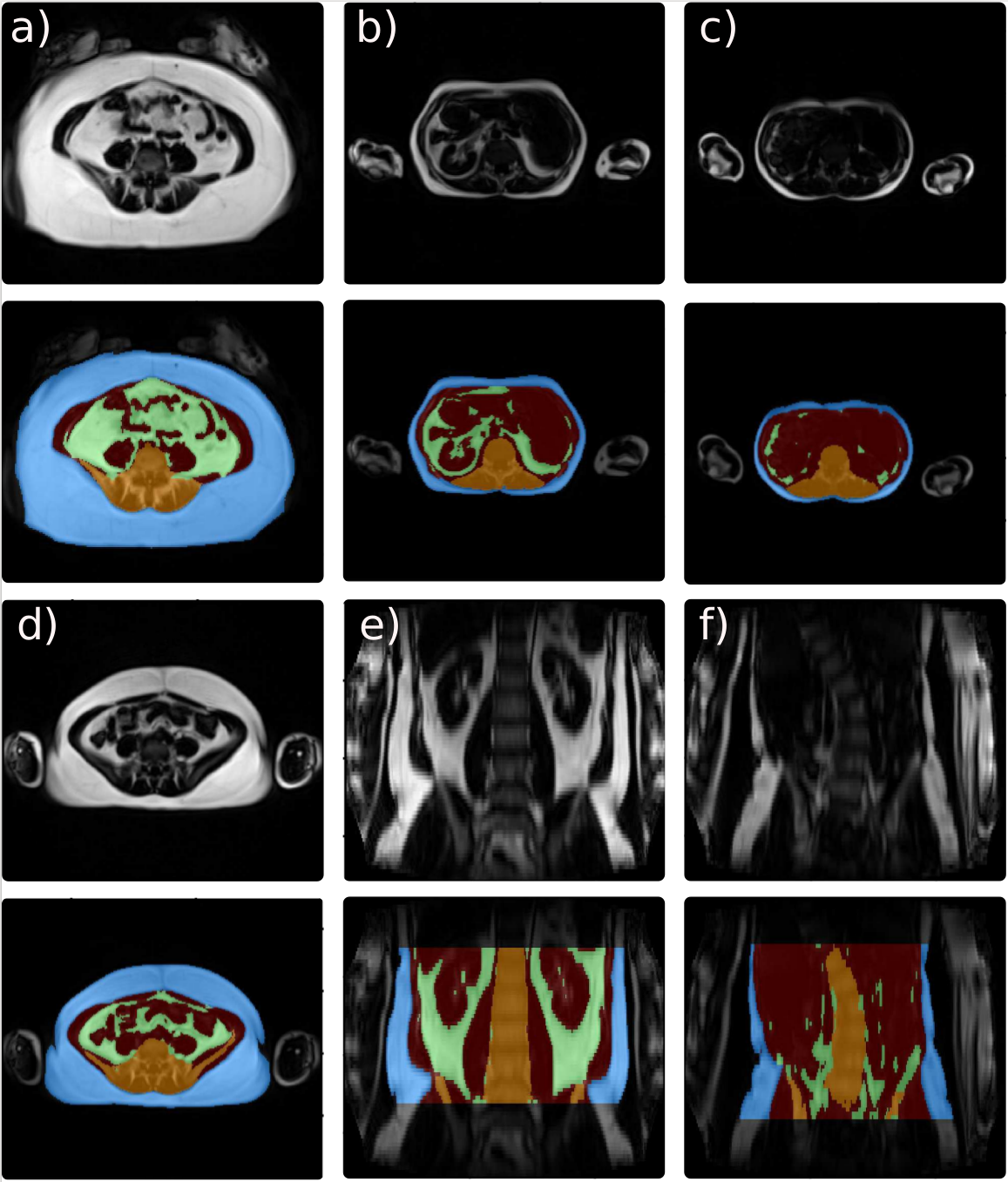}
\caption{\textbf{Examples of FatSegNet predictions on the Rhineland Study}. (a-f) Accurate automatic segmentation of different body shapes. Extreme cases: a) arms are in front of the abdominal cavity,and f) deviated spine.}
\label{fig:unseen_prediction_results}
\end{figure*}

For the first stage of the pipeline, i.e.\ localization of the abdominal region, all FCNNs can successfully determine the upper and lower limit of the abdominal region from a segmentation prediction map. However, our CDFNet requires significantly fewer parameters compared to the traditional UNet and Dense-UNet. Furthermore, the localization block is able to identify the abdominal region correctly even in cases with scoliosis (curved spine) as illustrated in Fig.~\ref{fig:unseen_prediction_results} f).
For the more challenging task of segmenting AAT, we demonstrate that CDF\-Net recovers VAT significantly better than traditional deep learning variants that rely on concatenation layers. Additionally, each individual CDF\-Net view model outperforms manual raters for segmenting the complex VAT and accomplishes equivalent results on SAT. The selection of an inhomogeneous BMI testing set ensures that our method is evaluated for different body types and avoids biases, as better segmentation performance can be achieved on subjects with high content of AAT compared to lean subjects~\cite{kullberg2007automated,addeman2015validation}. Moreover, images from individuals with high AAT could be accompanied by other types of issues, such as fat shadowing (Fig.~\ref{fig:unseen_prediction_results} d), or arms located in close proximity to the abdominal cavity (Fig.~\ref{fig:unseen_prediction_results} a), d) and e). These issues are mitigated by our view-aggregation model that regularizes the predicted segmentation by combining the spatial context from different views ultimately improving segmentation of tissue boundaries. Moreover, this approach automatically prevents misclassification of arms whereas previous deep learning AAT segmentation methods required manual removal of the upper extremities in a pre-processing step~\cite{adiposefully}. 
Note, that we prefer the 2D over a full 3D approach in this work. A full 3D network architecture has more parameters, requiring significantly more expert annotated training data (full 3D cases) and/or artificial data augmentation, which could increase the chance of overfitting -- in addition to increased GPU memory requirements.

 As demonstrated on the Rhineland Study data, the proposed pipeline exhibits high robustness and generalizability across a wide range of age, BMI, and a variety of body shapes as seen in Fig.~\ref{fig:unseen_prediction_results} and Fig.~\ref{fig:results_supp} a) and b). FatSegNet successfully identifies the AAT in different abdomen morphologies, spine curvatures, adipose shadowing, arms positioning, or intensity inhomogeneities. Furthermore, the pipeline has a high test-retest reliability between the calculated volumes of VAT and SAT without the need of any image pre-processing (bias-correction, image registration, etc.) or manual selection of a slice or region. Furthermore, the manually edited test set demonstrates a high similarity of automated and manual labels and excellent agreement of volume estimates.  However, as is usual with any automated method, segmentation reliability decreases when input images have low quality as illustrated in  Fig.~\ref{fig:results_supp} c), and d) where the scans present severe motion/breathing artifacts or very low image contrast. In order to detect these problematic images in large studies, an automated or manual quality control protocol should be implemented before passing images to automated pipelines.

 In accordance with previous studies on smaller data sets~\cite{sadananthan2015automated,sun2016automated}, our data showed a lower correlation of BMI with VAT-V than with AAT-V and SAT-V. We also observed a sex difference of the SAT-V and VAT-V accumulation as previously reported~\cite{machann2005age,kuk2005waist}: men were more likely to have higher VAT-V and lower SAT-V compared to women. Moreover, we further explored the association between age with SAT-V and VAT-V and found an obvious age effect on the accumulation of VAT-V in both men and women, and a weak age effect on SAT-V in women but not in men. This discrepancy was previously observed by Machann~\textit{et.al.}~\cite{machann2005age}, who assessed the body composition using MRI in 150 heal\-thy volunteers aged 19 to 69 years. They reported a strong correlation between VAT-V and age both in men and women, whereas SAT-V only slightly increased with age in women. The fact that our results replicate these previous findings on a large unseen dataset corroborates stability and sensitivity of our pipeline.
 
In conclusion, we have developed a fully automated post-processing pipeline for adipose tissue segmentation on abdominal Dixon MRI based on deep learning methods. While reducing the number of required parameters, the pipeline outperforms other deep learning architectures and demonstrates high reliability. Furthermore, the proposed method was successfully deployed in a large population-based cohort, where it replicated well known SAT-V and VAT-V age and sex associations and demonstrated generalizability across a large range of anatomical differences, both with respect to body shape and fat distribution.

\section{acknowledgements}
We would like to thank the Rhineland Study group for supporting the data acquisition and management, as well as Mohammad Shahid for his support on the deployment of the FatSegNet method into the Rhineland Study processing pipeline. Furthermore we acknowledge Tony St\"ocker and his team for developing and implementing the Dixon MRI sequence used in this work. This work was supported by the Diet-Body-Brain Competence Cluster in Nutrition Research funded by the Federal Ministry of Education and Research (BMBF), Germany (grant numbers 01EA1410C and FKZ: 01EA1\-809C), by the JPI HDHL on Biomarkers for Nutrition and Health (HEALTHMARK), BMBF (grant number 01EA1705B), the NIH R01NS0\-83534, R01\-LM012719, and an NVIDIA Hardware Award.

\clearpage
\section*{Supporting Information} 

\setcounter{figure}{0}
\renewcommand{\thefigure}{S\arabic{figure}}
\setcounter{table}{0}
\renewcommand{\thetable}{S\arabic{table}}

\begin{figure}[!hbt]
\centering
\caption{\textbf{View-aggregation Network} The proposed network is composed of a initial 3D convolution layer with 30 channels, followed by a batch normalization and a 3D convolutional layer for reducing the feature map dimensionality into the number of classes(n=5).}
\label{fig:view_Agg}
\end{figure}

\begin{figure}[!hbt]
\centering
\caption{\textbf{Step 1: Abdominal region localization.} Dice scores box-plot:  Average Dice score (cross-validation) of the abdominal region detection comparing the Proposed CDFNet vs. other FCNN architectures. The Dice scores are calculated on the average abdominal region generated from the average bounding boxes of the sagittal and coronal model. There is no significant difference between models, nonetheless, the proposed method achieves the same performance with $\sim$30~\% and $\sim$80~\% less parameters compared to Dense-UNet and UNet, respectively.}
\label{fig:loc_boxplot}
\end{figure}

\begin{figure}[!hbt]
\centering
\caption{\textbf{Comparison of single view model (left) vs. view-aggregation (right)}:
AAT predictions of two unseen subjects: a) normal subject, b) obese subject. View-aggregation avoids arm-misclassification (red boxes) and improves SAT (purple box).}
\label{fig:arms}
\end{figure}

\begin{table}[!hbt]
\centering
\caption{ \textbf{Case study analysis on the Rhineland Study data }. Characteristics of the  participants (n=587) showing mean (SD) for continuous and counts (PCT) for categorical variables}
\label{table:population}
\end{table}

\end{document}